\PassOptionsToPackage{numbers}{natbib}
\documentclass{article}

\usepackage[final]{nips_2017}

\usepackage[pointedenum]{paralist}
\usepackage{epic,eepic,eepicemu}

\usepackage[utf8]{inputenc} 
\usepackage{hyperref}       
\usepackage{url}            
\usepackage{booktabs}       
\usepackage{amsfonts}       
\usepackage{nicefrac}       
\usepackage{microtype}      
\usepackage{listings}

\usepackage{times}
\usepackage{graphicx,subfig}
\usepackage{graphics}
\usepackage{tabularx}
\usepackage{multirow}
\usepackage{amsfonts,amsmath,amssymb,amsthm,url,xspace}

\usepackage{paralist}
\usepackage[para,hang,flushmargin]{footmisc}
\usepackage{tablefootnote}

\usepackage[ruled,vlined, linesnumbered]{algorithm2e}
\usepackage{enumitem}

\newcommand{\R}{\mathcal{R}}
\newcommand{\F}{\mathcal{F}}

\renewcommand{\R}{\mathbb{R}}

\newcommand{\bx}{{\boldsymbol{x}}}

\def\dataepsilon{{\sf epsilon}\xspace}
\def\yahooltr{{\sf Yahoo-LTR}\xspace}
\def\msltr{{\sf Microsoft-LTR}\xspace}
\def\bosch{{\sf Bosch}\xspace}
\def\expo{{\sf Expo}\xspace}
\def\higgs{{\sf Higgs}\xspace}

\makeatletter
\renewcommand{\@noticestring}{}
\makeatother

\newcommand\Mark[1]{\textsuperscript{#1}}

\title{GPU-acceleration for Large-scale Tree Boosting}

\author{
  Huan Zhang\Mark{$\dagger$} \hspace{2.5em} Si Si\Mark{$\ddagger$} \hspace{2.5em} Cho-Jui Hsieh\Mark{$\star$} \\ 
  \Mark{$\dagger$}Dept. of Electrical and Computer Engineering, University of California, Davis\\
  \Mark{$\ddagger$}Google Research,
  Mountain View, CA\\
  \Mark{$\star$}Depts. of Statistics and Computer Science, University of California, Davis\\
  \texttt{ecezhang@ucdavis.edu}, 
  \texttt{sisidaisy@google.com},
  \texttt{chohsieh@ucdavis.edu}\\
}

\begin{document}

\maketitle

\begin{abstract}
 In this paper, we present a novel massively parallel algorithm for accelerating the decision tree building procedure on GPUs (Graphics Processing Units), which is a crucial step in Gradient Boosted Decision Tree (GBDT) and random forests training. Previous GPU based tree building algorithms are based on parallel multi-scan or radix sort to find the exact tree split, and thus suffer from scalability and performance issues. We show that using a histogram based algorithm to approximately find the best split is more efficient and scalable on GPU.  By identifying the difference between classical GPU-based image histogram construction and the feature histogram construction in decision tree training, we develop a fast feature histogram building kernel on GPU with carefully designed computational and memory access sequence to reduce atomic update conflict and maximize GPU utilization. Our algorithm can be used as a drop-in replacement for histogram construction in popular tree boosting systems to improve their scalability. As an example, to train GBDT on \dataepsilon dataset, our method using a main-stream GPU is 7-8 times faster than histogram based algorithm on CPU in LightGBM and 25 times faster than the exact-split finding algorithm in XGBoost on a dual-socket 28-core Xeon server, while achieving similar prediction accuracy.
\end{abstract}

\vspace{-5pt}\section{Introduction}
\vspace{-5pt}

Decision tree ensemble algorithms are increasingly adopted as a crucial solution to modern machine learning applications such as ranking and classification. 
The major computation cost of training decision tree ensemble comes from training a single decision tree, and the key challenge of decision tree building process is the high cost in finding the best split for each leaf, which requires scanning through all the training data in the current sub-tree. 
Since a tree ensemble algorithm typically has more than a hundred trees, while each tree has around 10 layers, the computation may require thousands
of data passes. As a result, training tree ensemble algorithms is time consuming for datasets with millions of data points and thousands of features.

Several parallel algorithms have been proposed to solve the scalability issue of building decision trees in multi-core or distributed settings. For example, 
XGBoost~\cite{chen2016xgboost} has a high-quality multi-core implementation for gradient boosted decision tree (GBDT) training which partitions the work by features, and it has been extended to distributed settings like Hadoop or Spark.
Recent works~\cite{FA16a,QM16a} proposed several different approaches for parallelizing
decision tree building on distributed systems, and 
LightGBM\footnote{{https://github.com/Microsoft/LightGBM}}, a popular tree boosting software package considers both data-partitioning and feature partitioning in their distributed version. 

As an important resource of parallel computing, GPU is much cheaper than building distributed systems and becomes a standard computing unit for big data analytics. 
However, the use of GPU is seldom exploited in decision tree building process and tree ensemble algorithms. 
To the best of our knowledge, among popular decision tree implementation packages, only XGBoost implements a GPU accelerated algorithm, but we find that it is slower than running on a 28-core CPU in our benchmarks because it implements the exact-split algorithm on GPU. Furthermore, due to GPU's strict memory constraint this method does not scale to large dataset. 

In this paper, we propose a novel GPU-based algorithm for building decision tree using a histogram-based method, and observe a significant speedup over all the existing multi-core CPU and GPU implementations for GBDT training. 
Unlike previous GPU approaches that parallelize sorting and scanning to find the exact split, our algorithm constructs histograms for all features on GPU, and then approximately finds the best split using these histograms. Our main contribution is in three folds:
\begin{itemize}[leftmargin=*]
    \item We show that histogram based methods for decision tree construction on GPU is more efficient than existing approaches, 
    which are based on multi-scan and radix sort to find the exact split. The exact-split based GPU tree builder in XGBoost actually cannot compete with a server with 28 cores. We design a very efficient algorithm for building feature histograms on GPU and integrate it into a popular GBDT learning system, LightGBM.
    \item We show significant speedup on large-scale experiments. For \dataepsilon dataset, XGBoost (with exact-split tree builder) takes over 4,100 seconds on a 28-core machine and we only need 165 seconds to achieve the same accuracy using a \$500 GPU, or 300 seconds with a \$230 GPU; comparing with histogram based method used in LightGBM, our GPU algorithm is also 7 to 8 times faster, while achieving a similar level of accuracy.
    \item Compared with existing GPU acceleration implementations for decision tree building, our scalability is much better. 
    The exact-split based GPU implementation in XGBoost fails due to insufficient memory on 4 out of 6 datasets we used, while our learning system can handle datasets over 25 times larger than \higgs on a single GPU, and can be trivially extended to multi-GPUs.
\end{itemize}


\vspace{-5pt}\section{Related Work}
\vspace{-5pt}
Decision tree has become one of the most successful nonlinear learning algorithms in many machine learning
and data mining tasks. Many algorithms are proposed based on decision trees and tree ensemble methods, such as random forest~\cite{Ho:1995:RDF:844379.844681,LB01a,RR02}, gradient boosting decision trees (GBDT)~\cite{JF01a}, 
and regularized greedy forest~\cite{johnson2014learning}. These algorithms have shown superb performance in regression, classification, and ranking~\cite{Li:2007} tasks. 

Among these tree ensemble methods, GBDT has gained lots of attention recently due to its superb performance and its flexibility of
incorporating different loss functions, such as square loss, logistic loss and ranking loss. 
Although the idea of GBDT is simple, it is actually non-trivial to have an implementation that performs well in practice, leading to a need of developing efficient and easy-to-use software
packages of GBDT. XGBoost~\cite{chen2016xgboost} is the most widely used package for training GBDT, and has shown lots of success in many data mining challenges. In terms of implementation, XGBoost uses several tricks: it uses the sort-and-scan algorithm discussed in Section~\ref{sec:DT} to find the exact best split on each leaf, designs regularization terms to prevent over-fitting, and
optimizes the code to handle different types of data and improve cache locality.
Recently, another GBDT package,  LightGBM
proposes to use histogram-building approach to 
speed up the leaf split procedure when training decision trees. Although the split of leaves is approximate, it is much more efficient than the exact-split method\footnote{XGBoost recently also added a histogram based learner}. We will discuss histogram based tree splitting in detail
in Section~\ref{sec:background-histogram}. 

As the size of data grows dramatically, there has been an increasing need for parallelizing decision tree training. The crucial part of decision tree training is to determine the best split of each leaf which turns out to be the main parallelizable component. One category of parallel decision tree algorithms is to partition the training data across machines, examples include PLANET~\cite{BP09a} and Parallel Voting Decision
Tree (PV-Tree) \cite{QM16a}.
They select top-k features within each machine, and then communicate to select the best split based on the feature histogram. 
Another group of approaches partition data by feature, and YGGDRASIL\cite{FA16a} is a representative of this category. 
The main idea of YGGDRASIL is to divide features into different machines, compute a local optimal split, and then master will decide the best split among them. 
XGBoost and LightGBM also have implemented distributed decision tree training implementation for both cases: partition over features and partition over data samples. 

The  previous works mainly focus on using CPU and multiple machines to parallelize decision tree training, however, as an important parallel computing resource, GPU is rarely exploited for this problem. Among these packages, only XGBoost utilizes GPU to accelerate decision tree training, but the speedup is not that significant, e.g., training on a top-tier Titan X GPU is only 20\% faster than a 24-core CPU\footnote{http://dmlc.ml/2016/12/14/GPU-accelerated-xgboost.html}.  There are also some other early attempts on building decision trees using GPUs, for instances, CUDATree\cite{liao2013learning}.
All these GPU implementations use a similar strategy to find the best split, which mimics the exact-split method on CPU. For the first a few tree levels, specialized multi-scan and multi-reduce operations are used to find splits among all leaves, and then it switches to radix sort and prefix-scan on re-partitioned data on each leaf when the tree goes deeper. It requires a lot of irregular memory access and its computation pattern does not fit into GPU's parallelization model well, so they can hardly compete with optimized multicore implementations on modern server CPUs.

\vspace{-5pt}\section{Problem Formulation and Background}
\label{sec:background}
\vspace{-5pt}
We first describe the standard procedure for training a decision tree, and then introduce a popular tree ensemble algorithm, Gradient Boosted Decision Tree (GBDT) that constructs a bunch of decision trees in a boosting fashion. After that, we will discuss the histogram based method to approximately construct decision trees, and how to efficiently use this method on GPU.


\vspace{-5pt}\subsection{Decision Tree}
\label{sec:DT}
\vspace{-5pt}
In this section we use regression tree to introduce the algorithm, while a similar approach can be used for classification trees. 
Given training data $X=\{\bx_i\}_{i=1}^N$ and their target $Y=\{y_i\}_{i=1}^N$, where $\bx_i\in\R^d$ and $\bx_i$ can be either continuous or categorical features. We use $X\in \R^{N\times d}$ to denote the data matrix, and $\bar{\bx}_j$ to denote the $j$-th column of $X$, which contains all the data points' value for $j$-th feature. A decision tree learns a model $f$ such that $f(\bx_i) \approx y_i$ with respect to some user defined loss functions. 
With square loss, the objective function for training a decision tree can be written as
\begin{equation}
    \min_{f\in\F} \sum\nolimits_{i=1}^N (f(\bx_i) - y_i)^2.
\label{eq:dt_loss}
\end{equation}
For illustration and simplicity, we use square loss throughout the paper and omit the discussion of the regularization term. We refer the readers to~\cite{JF01a,friedman2002stochastic} for more technical details on tree boosting.

The standard procedure to build a regression tree starts from the root node (containing all the training samples), and grows the tree by keeping splitting existing leaf nodes in the tree until some stopping conditions are met. Let $V_s$ denote the set of examples that pass through the leaf $s$ and define a split as $t = [\text{feature id}, \ \text{threshold}]$, consisting of the feature variable to split and at what threshold it has to be split. Based on the split, $V_s$ is partitioned into two disjoint sets: a set $V_r$ associated with the right node and a set $V_l$ associated with the left node. For each split pair we can compute the prediction values ($h_r$ and $h_l$) associated with the right and left nodes based on the loss function restricted to the corresponding sets of examples:
\begin{equation}
    \min_{h_l, h_r} \sum_{i\in V_r } (h_r-y_i)^2 + \sum_{i\in V_l } (h_l-y_i)^2, 
    \label{eq:loss_dtree}
\end{equation}
and the optimal assignment for $h_l$ and $h_r$ is 
\begin{equation}    
\label{eq:opt_dtree}
    h_l=(\sum\nolimits_{i\in V_l}y_i)/N_l,\\ \ \ \
    h_r=(\sum\nolimits_{i\in V_r}y_i)/N_r,
\end{equation}
where $N_l$ and $N_r$ are the number of data examples landed in left and right child respectively. After plugging~\eqref{eq:opt_dtree} into the loss function of ~\eqref{eq:loss_dtree}, the objective value for a given split $t$ becomes
\begin{equation}
   L(t) = \sum_{i}y_i^2- (\sum\nolimits_{i\in V_r}y_i)^2/N_r -  (\sum\nolimits_{i\in V_l}y_i)^2/N_l. 
\label{eq:obj_dtree}
\end{equation}

To find the best split, we need to test on all possible split pairs including all the feature id and their feature values and choose the one that achieves the lowest objective value in \eqref{eq:obj_dtree}. To reduce redundant computation, we first sort the $j$-th feature values for all examples on this leaf. Then, we
scan through the sorted feature values one by one to enumerate all possible split points, and at each step we move one example
from right child to the left. Assume knowing $\sum_{i}y_i$ (constant), then by maintaining the prefix sum $\sum_{i\in V_l} y_i$ when moving examples from right to left, 
the new loss~\eqref{eq:obj_dtree} can be computed in constant time at each step. After going over all examples on this leaf we find the exact feature value that minimizes~\eqref{eq:obj_dtree}.
In summary, there are two major computation steps in this exact-split finding method: (1) sort feature values; (2) update prefix sum $(\sum_{i\in V_r}y_i)$. All the previous attempts for GPU-based decision tree training rely on parallelizing these two steps. 

\vspace{-5pt}\subsection{Gradient Boosted Decision Tree}
\vspace{-5pt}
\label{sec:gbdt}
Gradient Boosted Decision Trees (GBDT) 
is a tree ensemble algorithm, that 
builds one regression tree at a time by fitting the residual of the trees that preceded it.
Mathematically, given a twice-differentiable loss function $\ell(y,X)$, GBDT minimizes the loss (for simplicity, we omit the regularization term here): 
\begin{equation}
L = \sum\nolimits_{i=1}^N \ell(y_i,F(\bx_i)), 
\end{equation}
with the function estimation $F(\bx)$ represented in an additive form:
\begin{equation}
 F(\bx) = \sum\nolimits_{m=1}^T f_m(\bx),
 \end{equation}
where each $f_m(\bx)$ is a regression tree and $T$ is the number of trees. GBDT learns these regression trees in an incremental way: at $m$-stage, fixing the previous $m-1$ trees when learning the $m$-th trees. More specifically, to construct the $m$-th tree, GDBT minimizes the following loss:
\begin{align}
L_m = \sum\nolimits_{i=1}^N \ell(y_i,F_{m-1}(\bx_i)+f_m(\bx_i)),
\label{eq:gb2}
\end{align}
where $F_{m-1}(\bx) = \sum_{k=1}^{m-1} f_k(\bx)$.
A popular way to solve the optimization problem in \eqref{eq:gb2} is by Taylor expansion of the loss function
\begin{align}
L_m\approx\bar{L}_m &= \sum_{i=1}^N
\big[
\ell(y_i,F_{m-1}(\bx_i))+g_if_m(\bx_i)+\frac{h_i}{2}f_m^2(\bx_i) \big],\\ \nonumber
\text{with}\ \ \ &g_i = \frac{\partial \ell(y_i,F(\bx_i))}{\partial F(\bx_i)}|_{F(\bx_i)=F_{m-1}(\bx_i)} \qquad
h_i = \frac{\partial^2 \ell(y_i,F(\bx_i))}{\partial^2 F(\bx_i)}|_{F(\bx_i)=F_{m-1}(\bx_i)} \nonumber
\end{align}
It is easy to see that minimizing $\bar{L}_m$ is equivalent to minimizing the following function: 
\begin{equation}
    \min_{f\in\F} \sum\nolimits_{i=1}^N \frac{h_i}{2} ( f_m(\bx_i) + \frac{g_i}{h_i})^2. 
    \label{eq:aabb}
\end{equation}

Interestingly, this optimization problem is equivalent to training a regression tree, as shown in~\eqref{eq:dt_loss}: 
Therefore, we can follow the similar procedure as discussed in \ref{sec:DT} to build a regression tree. 

\vspace{-0.6em}
\subsection{Approximate Split Finding Using Feature Histograms}
\vspace{-0.5em}
\label{sec:background-histogram}

As discussed in \ref{sec:DT}, finding the exact best split for a feature requires going through all feature values and evaluating objective function values for each of them. For large datasets, it is unnecessary and repetitious to check every possible position to find the exact split location; instead, an approximately best split often works quite well. 
One way to find the approximate best split is to test only $k$ split positions, and this can be done efficiently using feature histograms. We first convert continuous feature values into $k$ discrete bins, and then construct a histogram with $k$ bins for each feature. To find the split, we can  evaluate (\ref{eq:obj_dtree}) only at these $k$ points. Because building histograms is a rather straight-forward process, it is easy to implement efficiently on hardware. LightGBM and the ``hist'' tree builder in XGBoost use this approach to speed up decision tree training.

As we have shown in~\eqref{eq:aabb}, the objective function value for each decision tree in GBDT can be evaluated as long as we have the values of $g_i$ and $h_i$.
Therefore, the histogram-based algorithm usually build two histograms for $g_i$ and $h_i$, and then use them to find the best split. 
Algorithm~\ref{alg:feature-histogram} shows how to build a feature histogram. Note that we assume the feature values have been converted into $k$ integer bin values when data is loaded, thus $\bar\bx_j \in \{1, \cdots, k\}^N$. It is clear that this algorithm is memory-bound, as building the histograms requires non-sequential scattering access to large arrays. We utilize the high bandwidth memory and large computation power on GPUs to accelerate this operation.

\setlength{\textfloatsep}{5pt}
\begin{algorithm}
  \KwIn{$N_r$: number of data on leaf $r$ \\
  $h_i$: hessian of the loss on sample $i$ \\
  $g_i$: gradient of the loss on sample $i$ \\
  $d_{ri}$: $i$-th sample's index on leaf $r$, $|d_r|=N_r$ \\
  $\bar\bx_{ji}$: $i$-th value of binned feature $\bar\bx_j \in \{1,\cdots,k\}^N$}
  \KwOut{$o_{l,c}$: histogram for this leaf, with $l \in \{1,\dots,k\}$ are the $k$ bins, and $c \in \{1,2,3\}$ denote accumulated gradient, hessian and sample count respectively. }
  \For{$i \leftarrow 1$ \KwTo $N_r$} {
    $b \leftarrow \bar{\bx}_{j{d_{ri}}}$ \\
    $o_{b,1} \leftarrow o_{b,1} + g_{d_{ri}}$ \\ 
    $o_{b,2} \leftarrow o_{b,2} + h_{d_{ri}}$ \\
    $o_{b,3} \leftarrow o_{b,3} + 1$ \\
  }
  \caption{\small Construct feature $\bar\bx_j$'s histograms for leaf $r$}
  \label{alg:feature-histogram}
\end{algorithm}

\vspace{-0.8em}
\section{Proposed Algorithm}
\vspace{-0.5em}

\textbf{A Brief Review of Programming Challenges on GPUs.}
Although a GPU can easily handle a million threads, there are many constraints on how these threads can interact and work efficiently. First, a GPU executes a bundle of threads (called a ``warp'' or a ``wavefront'') in lock-step; in other words, threads within a bundle must be executing exactly the same sequence of instructions. Branching within a bundle will cause all threads in the bundle execute both directions of the branch, with certain threads masked off if the branching condition does not hold.
Moreover, unlike threads on a CPU, GPU threads cannot synchronize and communicate with each other in an arbitrary manner. Only a small group of threads (called a \textit{workgroup}, usually in size of a few hundreds), can synchronize and efficiently exchange data with each other. Each workgroup has access to a dedicated high-bandwidth and low latency \textit{local memory}, which can be used as a scratchpad or to exchange data between threads within a workgroup. Local memory usually has a very limited size (for example, $\leq 64$ KBytes per workgroup), and using too much local memory can affect performance adversely.

Although GPU's main memory (\textit{global} memory) can have 10 times more bandwidth than CPU's main memory, loading data from global memory can be quite expensive with a very long latency.
One remarkable feature of GPUs is that they can make context switches between thread bundles at very little cost, so when a bundle of threads are stalled due to global memory access, other available threads can be dispatched. As long as we have enough threads to schedule, the long memory latency can be hidden; thus it is important to occupy the GPU with a sufficiently large number of threads.

\textbf{Building Histograms on GPU: problem of having too many threads.}
\label{sec:gpu-image-histo}
Building image histograms using GPUs is a classical task in general-propose GPU computing~\cite{wilt2013cuda}. Given an image as an array, we maintain a counter for each bin, and increment the corresponding counter if a image pixel value falls into that bin. The single thread implementation of this algorithm is trivial; however, problems arise when there are a large number of threads computing one histogram. One way to build histogram in parallel is that each thread builds its private histogram using part of the data, preferably in its local memory, and in the end all threads reduce their private histograms into a single final histogram. 
When the number of threads is large, the reduction step will incur a large overhead; also, the limited size of local memory prevents us from building too many private histograms. Thus, we want the number of private histograms to be much smaller than the total number of threads. However, if two or more threads update the bin counters of the same histogram, their updates may conflict with each other, i.e., two or more of them may want to increment the same counter. In this case, we have to guarantee that when a conflict occurs, threads resolve it by updating the counter sequentially, one by one. To do this efficiently without explicit locking, hardware atomic operations are necessary. Fortunately, most recent GPUs (AMD GPUs past 2012 and NVIDIA GPUs past 2014) support hardware atomic operations in local memory~\cite{mantor2012amd,nvtunemaxwell}, but it is still important to reduce conflicts for best performance.

\textbf{Constructing Feature Histograms on GPU.}
When we build feature histograms for decision tree learning, our input is a set of features with corresponding statistics (gradient and hessian) rather than an image.
There are some important differences between building image histograms and feature histograms. First, unlike constructing the histogram for an image, building feature histograms involves building a large number of histograms at the same time, one for each feature. Second, besides incrementing an integer counter for each bin, we need to accumulate gradient and hessian statistics, two floating point numbers for each bin. Third, since we only need to access samples on the current leaf, the memory accessing pattern for loading each sample's feature is non-sequential. 

\textbf{Data Structure and Memory Allocation.}
Since we need non-sequential scatter access to the feature array and global memory access is expensive, it is very inefficient to just read one byte of feature data. Thus, we bundle every 4 binned features (one byte each) into a 4-feature tuple (4-byte) and store feature tuples in GPU memory. Each GPU thread will work on 4 features of one sample at once. Since GPU is built for single precision (4-byte) arithmetic, 4-byte elements usually yields best efficiency. This strategy also requires that each workgroup maintains 4 set of histograms in local memory, and each set of histogram consists of 3 statistics: gradient, hessian and a counter. Each value takes 4 bytes (assuming single precision is used), so the total local memory requirement is $4 \times 3 \times 4 \times k$ bytes. When $k=256$, we need 12 KB local memory per workgroup. This allows 5 workgroups per compute unit of GPU 
, which is an acceptable occupancy.

\textbf{Reduce Atomic Update Conflicts.}
For simplicity, here we focus on discussing how to build 8 histograms (gradient histogram and hessian histogram for 4 features) at the same time. In our GPU algorithm, each thread processes one sample of the 4-feature tuple, and a total of $m$ ($m$ is the size of workgroup) samples are being processed at once. Remember that GPUs execute a bundle of threads in lock-step. It is easy to write the program as every thread updates feature 0's gradient histogram all together, then updates feature 1's gradient histogram all together, etc, until all 4 features' gradient and hessian histograms are updated. The update operation needs to be atomic.
However, when all $m$ (usually 256) threads update a single histogram with $k$ bins simultaneously, it is very likely that some threads have to write to the same bin because they encounter the same feature value, and the atomic operation becomes a bottleneck since hardware must resolve this conflict by serializing the access. 

To reduce the chance of conflicting updates, we exploit a special structure that occurs in our feature histogram problem but not in traditional image histogram problem---we construct multiple histograms simultaneously instead of just one. We want $m$ threads to update all 8 distinct histograms in each step, as shown in Algorithm~\ref{alg:gpu-histogram}. To understand this algorithm, we can consider a special case where $m=8$. In line~\ref{alg-line:acc1}, when $l=0$, thread 0, 1, 2, 3 update the gradient histogram of feature 0, 1, 2, 3 using data sample $i=0, 1, 2, 3$'s feature value, while thread 4, 5, 6, 7 update the hessian histogram of feature 0, 1, 2, 3 using data sample $i=0, 1, 2, 3$'s feature value. In this case, $m$ threads are updating $8k$ histogram bins at each step, greatly reduce the chance that two threads write to the same bin. For real implementation, line~\ref{alg-line:acc1} and~\ref{alg-line:acc2} require some tricks because we must avoid the \texttt{If} statement.

\begin{algorithm}[t]
  \KwIn{$\bar\bx_{ji}$: $i$-th value of binned feature $\bar\bx_j \in \{1,\cdots,k\}^N$, $k$ is the bin size\\
  $N_r$: number of data on leaf $r$ \\
  $d_{ri}$: $i$-th sample's index on leaf $r$, $|d_r|=N_r$ \\
  $g_{ri}$: gradient of the $i$-th sample on leaf $r$\\ 
  $h_{ri}$: hessian of the $i$-th sample on leaf $r$
  }
  \textbf{All $m$ threads in this workgroup execute the following simultaneously, in lock-step:} \\
  tid $\leftarrow$ thread ID $\in \{0, 1, \dots,m-1\}$, $i \leftarrow$ tid \\
  $s \leftarrow$ bit 0 and 1 of tid, $c \leftarrow$ bit 2 of tid \\
  \While{$i < N_r$} {
    $t \leftarrow$ ($\bar\bx_{j,{d_{ri}}},\bar\bx_{j+1,{d_{ri}}},\bar\bx_{j+2,{d_{ri}}},\bar\bx_{j+3,{d_{ri}}}$) \\
    $g^\prime \leftarrow g_{ri}$, $h^\prime \leftarrow h_{ri}$ \\
    \For{$l \leftarrow 0$ \KwTo $3$} {
      $p \leftarrow (l+s) \mod 4$ \\
      $b \leftarrow $ the $p$-th feature value in 4-tuple $t$ \\
      every thread where $c=0$ accumulates $g^\prime$ to feature $p$'s gradient histogram atomically and every thread where $c=1$ accumulates $h^\prime$ to feature $p$'s hessian histogram atomically based on the bin value $b$\\
      \label{alg-line:acc1}
      every thread where $c=0$ accumulates $h^\prime$ to feature $p$'s hessian histogram atomically and every thread where $c=1$ accumulates $g^\prime$ to feature $p$'s gradient histogram atomically based on the bin value $b$\\
      \label{alg-line:acc2}
    }
    $i \leftarrow i + m$ \\
  }
  Copy 4 sets of histograms in local memory to global memory \\
  \caption{\small Fast Feature Histogram Construction on GPU for a 4-feature tuple $(\bar\bx_j, \bar\bx_{j+1}, \bar\bx_{j+2}, ,\bar\bx_{j+3})$.}
  \label{alg:gpu-histogram}
\end{algorithm}

\textbf{Parallel By Features and Data.}
For simplicity, in Algorithm~\ref{alg:gpu-histogram} we only show the case where a 4-feature tuple is entirely processed by this workgroup, and this will require $\lceil d/4 \rceil$ workgroups to process $d$ features. In our implementation, we split the work to GPU workgroups both by features and by samples. If more than one workgroup is processing the same 4 features, a final reduction program (also runs on GPU) is required to merge their private histograms into the final histogram.

\textbf{Use of Small Bin Size.}
A major benefit of using GPU is that we can use a less than 256 bin size to further speedup training, potentially without losing accuracy. On CPU it is not very beneficial to reduce the bin size below 256, as at least one byte of storage is needed for each feature value.
However, in our GPU algorithm, using a smaller bin size, for example, 64, allows us to either add more private histograms per workgroup to reduce conflict writes in atomic operations, or reduce local memory usage so that more workgroups can be scheduled to the GPU, which helps to hide the expensive memory access latency. Further more, using a smaller bin size can reduce the size of histograms and data transfer overhead between CPU and GPU. We observe significant performance gain by using a bin size of 64, without losing training accuracy, as we will show in section~\ref{sec:exp}.

\section{Experimental Results}
\vspace{-0.5em}
\label{sec:exp}

We compare the following algorithms for decision tree learning in GBDT in this section:

\textbf{\sf Histogram}: 
We will compare our proposed histogram-based algorithm on GPU\footnote{Our GPU algorithm was first released at \url{https://github.com/huanzhang12/lightgbm-gpu} on Feb 28, 2017, and has been merged into LightGBM repository on April 9, 2017, in commit \texttt{0bb4a82}} with other methods.
LightGBM is a representative CPU implementation of this algorithm and we use it as the reference\footnote{XGBoost recently implemented {\sf Histogram} based tree construction using a similar algorithm as LightGBM}. 

\textbf{\sf Exact}: the traditional way to learn a decision tree as described in section~\ref{sec:DT}, which enumerates all possible leaf split points. We use the ``exact'' tree learner in XGBoost for the implementation on CPU, and the ``grow\_gpu'' learner~\cite{mitchell2017accelerating} in XGBoost on GPU as the reference implementation of {\sf Exact}~\footnote{\mbox{XGBoost added a {\sf Histogram} based GPU learner on April 25, 2017 with a similar basic algorithm as ours}}.

\textbf{\sf Sketching}: proposed in~\cite{chen2016xgboost}, which also uses histogram for approximately finding the split, however features are re-binned after each split using sketching. We use the ``approx'' tree learner in XGBoost for this algorithm. No GPU implementation is available for this algorithm due to its complexity.

\textbf{Datasets.} We use the following six datasets in our experiments: \higgs, \dataepsilon
,
\bosch
, \yahooltr
, \msltr, and \expo
, as shown in Table~\ref{tb:dataset}. The six datasets represent quite different data characteristics. The first three are very large and dense datasets collected for classification tasks; \yahooltr and \msltr are for learning to rank tasks with a mixture of dense and sparse features. The \expo dataset has categorical features. For comparisons involving XGBoost, we did not include the \expo dataset, as it does not support categorical features directly and 
has to convert the dataset to one-hot encoding, which makes the comparison unfair. 

\begin{table}[t]
\caption{\small Dataset statistics}
\centering
\resizebox{13cm}{!}
{
 \begin{tabular}{| c | c | c | c | c | c | c |} 
 \hline
  Datasets & \higgs\cite{baldi2014searching} & \msltr\cite{QinL13} & \yahooltr
  \tablefootnote{{https://webscope.sandbox.yahoo.com/catalog.php?datatype=c}} 
  & \dataepsilon
  \tablefootnote{{http://largescale.ml.tu-berlin.de}} 
  & \expo
  \tablefootnote{{http://stat-computing.org/dataexpo/2009}} 
  & \bosch
  \tablefootnote{https://www.kaggle.com/c/bosch-production-line-performance} 
  \\
 \hline
 Training examples & 10,000,000 & 2,270,296 & 473,134 & 400,000 & 10,000,000 & 1,000,000 \\
  \hline
 Feature Dimension & 28 & 137 & 700 & 2,000 & 700 & 968 \\
 \hline
\end{tabular}
}
\label{tb:dataset}
\end{table}

\textbf{Parameters.} We follow a publicly available benchmark instruction\footnote{https://github.com/Microsoft/LightGBM/wiki/Experiments} for setting training parameters, so that our results are comparable to public results. For {\sf Histogram} in LightGBM, we set the total number of leaves to 255, and each leaf has at least one example. Bin size $k$ is set to 255\footnote{LightGBM uses one bin as sentinel, thus a byte can only represent 255 bins. Similarly, only 63 bins are used when using a 6-bit bin value representation.} and 63. For {\sf Exact} and {\sf Sketching} in XGBoost, we set the maximum tree depth to be 8. The GPU implementation of {\sf Exact} algorithm in XGBoost only works on \bosch and \yahooltr datasets; other datasets do not fit into the 8 GB GPU memory. For all experiments, we use learning rate $\eta = 0.1$ and run 500 boosting iterations, except for \bosch we set $\eta = 0.015$.

\textbf{Hardware.}
In all our experiments, we use two representative, main-stream GPUs from the latest production line of AMD and NVIDIA: Radeon RX 480 and GTX 1080. The two GPUs are installed to a dual-socket 28-core Xeon E5-2683 v3 server with 192 GB memory, and we use the same machine to collect results for the CPU algorithms. For all CPU results, we run 28 threads. We list the characteristics of these hardware in Table~\ref{tb:hardware}. Note that the GPUs we used are not the best ones in the market, and our results can be further improved by using a more expensive GPU. We hope that even a budget GPU can show significant speedup in training, making GPU a cost-effective solution. 

\begin{table*}[htbp]
\caption{\small Hardware characteristics. We compare the performance of a 28-core server with a budget GPU (RX 480) and a main-stream GPU (GTX 1080). MSRP = Manufacturer's Suggested Retail Price.
}
\resizebox{14cm}{!}
{
 \begin{tabular}{| c | c | c | c | c |} 
 \hline
 Hardware & Peak FLOPS & Memory Size & Peak Memory Bandwidth & Cost (MSRP) \\ [0.5ex]
 \hline
 AMD Radeon RX 480 & 5,161 GFLOPS & 8 GB & 256 GB/s & \$239 \\
 \hline
NVIDIA GTX 1080 & 8,228 GFLOPS & 8 GB & 320 GB/s & \$499 \\
 \hline
 2x Xeon E5-2683v3 & 1,792 GFLOPS & 192 GB & 133 GB/s & \$3692 (CPU only) \\ 
 \hline
\end{tabular}
}
\label{tb:hardware}
\vspace{-0.5em}
\end{table*}

\textbf{Memory Usage Comparison.} 
As shown in Table~\ref{tb:memory}, our histogram-based method uses at most 1~GB GPU memory for all datasets, thanks to the fact that each feature value after binning only takes 1 byte. A GPU with 16 GB memory can deal with datasets at least 16 times larger than \dataepsilon, or over 25 times larger than \higgs. For even larger datasets, we can trivially extend our algorithm to multi-GPU cases where each GPU holds a disjoint set of features. As a comparison, we also include the GPU memory usage of {\sf Exact}. Unfortunately, it can easily run out of memory because GPU memory is usually much smaller than CPU. Considering most GPUs have 8 GB to 16 GB memory, {\sf Exact} algorithm on GPU requires too much memory for training most large scale datasets. Since GPU is particularly useful for training large datasets, the usefulness of {\sf Exact} on GPU is very limited.

\begin{table}[t]
\centering
\caption{\small GPU memory usage comparison between exact-split and our histogram-based tree split. Our GPU has 8 GB memory (typical for recent GPUs). OOM = out of memory. Unlike CPUs, GPUs typically have limited and unexpandable memory, thus it is important to use memory efficiently to scale to large datasets.}
\resizebox{13cm}{!}
{
 \begin{tabular}{| c | c | c | c | c | c | c |} 
 \hline
 Datasets & \higgs & \msltr & \yahooltr & \dataepsilon & \expo & \bosch \\
 \hline
 {\sf Exact} on GPU & OOM & OOM & 4165 MB & OOM & OOM & 7139 MB \\
 \hline
 {\sf Histogram} on GPU (ours) & 611 MB & 413 MB & 291 MB & 901 MB & 405 MB & 1067 MB \\
 \hline
\end{tabular}
}
\label{tb:memory}
\end{table}

\textbf{Training Performance Metrics Comparison.}
Since we use reduced precision and less number of bins for training on GPU, it is interesting to see whether training on GPU can obtain the similar level of performance metrics (AUC, NDCG) with training on CPU. In Table~\ref{tb:train-acc}, we can see that training with a bin size of 64 does not affect training performance metrics on both CPU and GPU. Also, our {\sf Histogram} based method on GPU can get very similar AUC and NDCG with the one on CPU despite using single precision. This table, on the other hand, justifies the use of a smaller bin size.

\begin{table*}[htbp]
\vspace{-0.5em}
\caption{\small Training metrics comparison between different algorithms. OOM = out of memory
}
\resizebox{14.5cm}{!}
{
 \begin{tabular}{| c | c | c | c | c | c | c | c | c |} 
 \hline
 Datasets & Metrics & \shortstack{{\sf Histogram}, \\ $k=255$, \\ CPU} & 
 \shortstack{{\sf Histogram} \\ $k=63$ \\ CPU} & 
 \shortstack{{\sf Histogram} \\ $k=255$ \\ GPU (ours)} & 
 \shortstack{{\sf Histogram} \\ $k=63$ \\ GPU (ours)} & 
 \shortstack{{\sf Exact} \\ CPU} & 
 \shortstack{{\sf Sketching} \\ CPU} & 
 \shortstack{{\sf Exact} \\ GPU} \\ 
 \hline
\higgs & AUC & 0.845612 & 0.845239 & 0.845612 & 0.845209 & 0.839593 & 0.841218 & N/A \\ \hline
\multirow{4}{*}{\msltr} & NDCG@1 & 0.521265 & 0.521392 & 0.521789 & 0.522163 & 0.480545 & 0.485436 & OOM \\
 & NDCG@3 & 0.503153 & 0.505753 & 0.503886 & 0.504089 & 0.465276 & 0.468778 & OOM \\
 & NDCG@5 & 0.509236 & 0.510391 & 0.509861 & 0.510095 & 0.471930 & 0.472537 & OOM \\
 & NDCG@10 & 0.527835 & 0.527304 & 0.528009 & 0.527059 & 0.491390 & 0.491403 & OOM \\ \hline
\multirow{4}{*}{\yahooltr} & NDCG@1 & 0.730824 & 0.730165 & 0.730936 & 0.732257 & 0.720044 & 0.720101 & 0.721566 \\
 & NDCG@3 & 0.738687 & 0.737243 & 0.73698 & 0.739474 & 0.717639 & 0.720497 & 0.718576 \\
 & NDCG@5 & 0.756609 & 0.755729 & 0.756206 & 0.757007 & 0.738333 & 0.739844 & 0.738822 \\
 & NDCG@10 & 0.79655 & 0.795827 & 0.795894 & 0.797302 & 0.781000 & 0.782008 & 0.781197 \\ \hline
\dataepsilon & AUC & 0.950243 & 0.949952 & 0.950057 & 0.949876 & 0.949725 & 0.949767 & OOM \\ \hline
\expo & AUC & 0.776217 & 0.771566 & 0.776285 & 0.77098 & N/A & N/A & OOM \\ \hline
\bosch & AUC & 0.718115 & 0.721791 & 0.717184 & 0.724761 & 0.700607 & 0.699601 & 0.705512\\ \hline
\end{tabular}
}

\label{tb:train-acc}
\vspace{-0.5em}
\end{table*}

\begin{figure*}[htbp]
\hspace{-2em}\includegraphics[width=0.7\paperwidth]{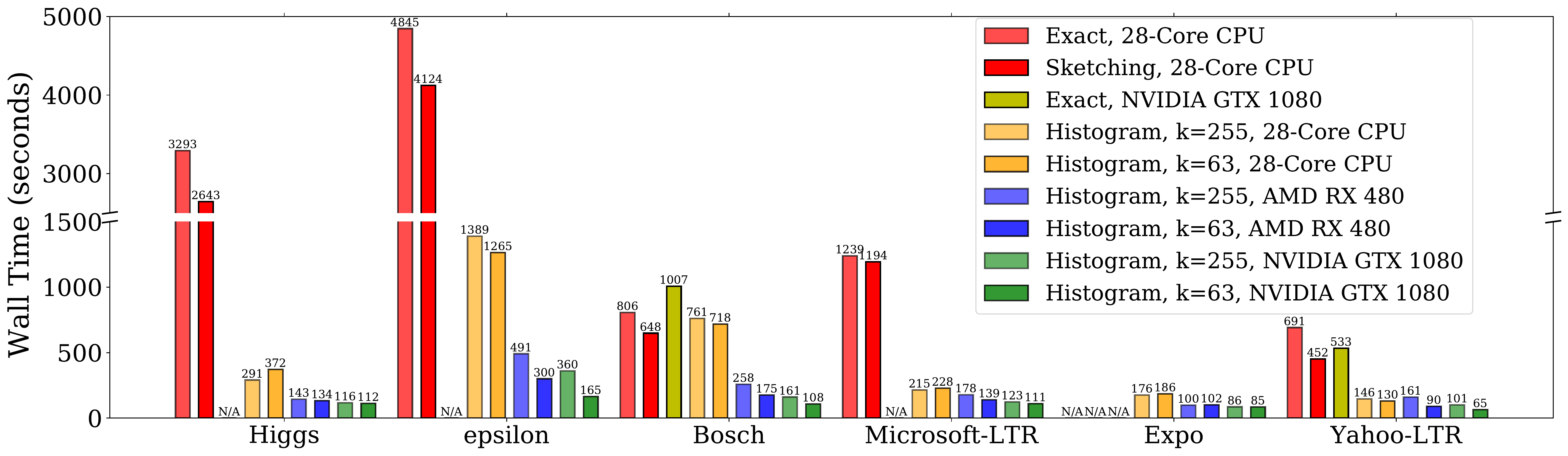}
\vspace{-1em}
\caption{\small Performance comparison between {\sf Histogram} with 63 and 255 bins on CPU and GPU, {\sf Exact} on CPU and GPU, and {\sf Sketching} on CPU. {\sf Exact} on GPU runs out of memory except for \bosch and \yahooltr}
\label{fig:training-speed}
\vspace{-0.5em}
\end{figure*}

\textbf{Training Speed.}
As shown in Figure~\ref{fig:training-speed}, on dataset \dataepsilon and \bosch, our speedup is most significant: Using the GTX 1080 GPU and 63 bins, we are  7-8 times faster than {\sf Histogram} algorithm on CPU, and up to 25 times faster than {\sf Exact} on CPU. On \higgs, \expo and \msltr, we also have about 2-3 times speedup. Even using a low-cost RX 480 GPU (less than half of the price of GTX 1080), we can still gain significant amount of speed up, as it is only 30\% to 50\% slower than GTX 1080. We should reemphasize that this comparison is made between a powerful 28-core server, and a budget or main-stream (not the best) GPU. 
Also, {\sf Exact} on GPU cannot even beat 28 CPU cores on \yahooltr and \bosch, and for all other datasets it runs out of memory.
Thus, we believe that the {\sf Exact} decision tree construction algorithm using parallel multi-scan and radix sort on GPU does not scale well. Our histogram based approach can utilize the computation power of GPU much better.

We encourage the reader to read the appendix for more design details and experimental results.

\medskip
\small

\bibliographystyle{unsrt}
\bibliography{main}

\newpage

\appendix

\section{Additional Design Considerations}

\subsection{Further Reduce Atomic Update Conflicts Using Banked Histogram Counters}

Along with the techniques we described in section 4 to reduce atomic update conflicts and increase performance, we also used banked histogram counters which is a traditional technique to speedup image histogram building on GPU. Instead of building just one histogram per feature that is shared by all threads, we build $B$ banks of histograms for each feature, and each thread in a workgroup only updates one of the counter banks. Finally, counter values of all banks will be summed up to one value. In this way, the conflict rate on updates is reduced by a factor of $B$.
However this option is severally limited to feature histogram building because of the limited local memory space. With bin size 256, feature histogram takes 12~KBytes local memory per workgroup and we do not use any additional bank. With bin size 64, feature histogram takes only 3~KBytes and thus we can afford 4 banks, and still use 12~KBytes per workgroup. This greatly improves performance for some datasets, as we have shown in section 5.

\subsection{Sparse Features}

When GPU is constructing histograms, we also want the CPU to do some useful work at the same time to maximize resource utilization. Currently, our strategy is to process sparse features on CPU and process dense features on GPU. Processing sparse features requires more bookkeeping during leaf split and histogram construction, thus it is beneficial to treat a feature as sparse only when it is sparse enough. By default, if a feature has more than 80\% zeros, LightGBM will treat it as a sparse feature and use special data structure to process it. We change this compile time threshold constant to a new configuration variable \texttt{sparse\_threshold} (denoted as $t$, with a default value of 0.8). LightGBM will process a feature as sparse only when there are more than $t \times N$ zeros. By selecting an appropriate threshold, we can balance the load on both computing resources and maximize performance.

In our experiments, we found that for \msltr and \yahooltr, it is best to make features completely dense to process them on GPU ($t=1$), because our GPU histogram construction algorithm has a large speedup factor. For datasets \higgs, \bosch, \dataepsilon and \expo, all features are already dense and processed on GPU.

\subsection{Bin Redistribution}

In some cases, only a few distinct values appear in a binned feature (for example, a feature only contains numbers drawn from $\{1, 2, 3\}$), thus the effective bin size will be set to 3 instead of specified 64 or 256. Although this is not a problem for CPU, on GPU this causes performance degradation as only a few bins are being written and the conflict rate during atomic operations will be high. Because threads within a bundle (``warp'') execute in locksteps, this can slow down an entire bundle of threads. Our benchmark shows that in the extreme case, where there exists one feature with only one bin value such that every thread contents to update the same bin, the workgroup (assuming a bin size of 256) processing that 4-feature tuple can be 4 times slower comparing with random feature values.

We solve this problem by redistributing the bin values. Before copying features to GPU, if we find that one feature only lies in $k^\prime < k/2$ bins, where $k$ is the desired maximum bin size for the dataset, we will redistribute the feature bin value with at most $k^\prime$ numbers using the strategy shown in Algorithm~\ref{alg:bin-redist}.

\begin{algorithm}
  $m \leftarrow 2^{\lfloor \log_2 \frac{k}{k^\prime} \rfloor}$ \\
  mask $\leftarrow m - 1$ \\
  \For{each feature bin value $\bar\bx_{ji}$} {
    $\bar\bx_{ji} := \bar\bx_{ji} \times m + (i \quad \textbf{bitwise\_and} \quad \text{mask})$
  }
  \caption{Bin Redistribution for feature $\bar\bx_j$}
  \label{alg:bin-redist}
\end{algorithm}

After we transfer the constructed histogram from GPU back to CPU, we will simply accumulate the values from bin $i \times m$ to $(i + 1) \times m - 1$ to get values of the $i$-th bin of the original histogram.

\subsection{Dealing With Repeated Feature Values}

In some datasets, we found that some binned features range from 1 to $k$ but most of them are just a single value.
It occurs quite often when a features has some sparsity and thus there are many repeated zeros. This can slow down the histogram construction as there are a lot update conflicts for bin 0. We add a fast path in our implementation for this case: we maintain the most recently used bin of the histogram counters in GPU registers, and when the same bin value occurs again we don't need to update the histogram in local memory.
Our benchmark shows that it incurs very little overhead, but noticeably decreases the training time of \yahooltr and \msltr.

\subsection{Data Movement}

Before we launch the GPU program to construct feature histograms, we need to prepare for its input. All feature values are copied as 4-feature tuples only once before training starts, and stay unchanged during the entire training process. After one leaf is built, CPU updates the list of sample indices and the corresponding gradient and hessian statistics for samples on that leaf. 
Thus three arrays (indices, gradients, hessians) need to be copied to GPU after each split. We make the copy asynchronous when possible, to hide the data transfer time.

Instead of transferring all histograms to CPU memory after all of them are built in GPU memory, 
we start transferring the histogram for each feature back to CPU memory immediately after it has been built. 
This also helps hiding data transfer time from histogram computation, especially when the number of features is large. 

\subsection{Numerical Issues}
Limited precision training has been successfully used in training of deep neural networks~\cite{gupta2015deep}. Theoretical analysis for limited precision training is also available for some algorithms like SGD~\cite{de2015taming}. In histogram based decision tree training, the main numerical issue comes from the accumulation of gradient and hessian statistics of each bin. The current CPU implementation of lightGBM uses 32-bit floating point to store statistics, but the accumulation is done by 64-bit floating point arithmetic.  However, unlike CPUs, most GPUs are relatively weak to compute in double precision. For example, the NVIDIA Titan X Pascal has over 10 TFLOPS peak processing power for single precision operations, but this number becomes only 0.3 TFLOPS for double precision floating point operations. Moreover, double precision histograms also increase the memory pressure of local memory.
Thus, it is necessary to avoid double precision computation on GPUs in order to have good performance. In our implementation, we use 32-bit single precision numbers by default. 
We also provide a configuration parameter to switch our algorithm into double precision mode for the cases when users do have a GPU with good double precision processing power (like NVIDIA Tesla series), or want better arithmetic precision for reproducibility.

\section{Implementation Details}

\subsection{GPU Algorithm Implementation}
Our GPU algorithm for building feature histograms is implemented in OpenCL 1.2 and can target a large range of GPU devices from different vendors. We guarantee our implementation quality by using inline assembly when possible and check the compiler generated GPU assembly code manually, to ensure that our code runs with best efficiency.

\subsection{Integrating Our Algorithm Into LightGBM}

GPU code is well known for being tricky, unfriendly to ordinary programmers and hard to maintain. Instead of re-implementing the entire tree building logic of LightGBM on GPU, we only implement the procedure of building feature histograms, which is the most time-consuming operation in LightGBM. Thus, our GPU algorithm has only weak interactions with other parts of LightGBM, and the new learning system immediately gains all other good capabilities of LightGBM, without re-implementing these features on GPU. Thanks to our implementation's modularity, our accelerated histogram building algorithm also works for distributed GBDT building methods in LightGBM, thus we make very large scale distributed GPU training possible.

We replace the ConstructHistogram function in LightGBM with our GPU implementation, and add necessary code for GPU initialization and data movement to LightGBM. Our implementation is publicly available\footnote{\url{https://github.com/huanzhang12/lightgbm-gpu}} since Feb 28, 2017, and has been officially merged into LightGBM in commit \texttt{0bb4a82} on April 9, 2017.

\subsection{Atomic Operations on GPU}
Most modern GPU architectures (NVIDIA Maxwell or later, AMD Graphic Core Next 1.0 or later) support atomic operations in local memory space.
However, these atomic instructions only work on integer data types. 
Although NVIDIA provides an pseudo (PTX) instruction \texttt{atom.shared.add.f32} to perform atomic floating point addition in local memory, it does not translate to a single hardware instruction. Rather, the compiler will generate a small loop consisting of a floating point addition and an atomic compare-and-swap operation in local memory. On AMD GPUs, no such pseudo instruction is available, so we directly implement atomic floating point addition using a loop of local memory compare-and-swap instructions.

\subsection{Disabling Power Saving Features} Many GPUs have aggressive DVFS (dynamic voltage and frequency scaling) behaviour which downclocks the GPU when it is not under full load. Since our algorithm does not work on GPU with full load (CPU is used to finalize the tree with GPU constructed histograms), GPU drivers are likely to automatically select a lower power state with reduced frequency. To overcome this problem on AMD GPUs, we manually set performance mode using control knobs located at \texttt{/sys/class/drm/}.
For the NVIDIA GPU we used, we cannot find a reliable way to put it into the \texttt{P0} (highest) performance mode. Thus, our code runs in \texttt{P2} mode with reduced GPU memory clock. Even though, we are still able to show significant speedup.

\subsection{Turbo Boost and Hyper-threading} For better reproducibility, we disable Turbo Boost on CPU and run the CPU in performance mode (fixed maximum frequency). We also do not use hyper-threading, as we found that LightGBM becomes slower with hyper-threading in our case (with 28 CPU cores) due to additional threading overhead.

\section{Additional Experimental Results}

\subsection{Performance Characterization of Histogram-based Algorithm on CPU}

We make an instruction-level profiling of the {\sf Histogram} algorithm in LightGBM on CPU, and identified the major bottlenecks over several different datasets. We found that over 85\% of time is spent on the four functions in Table~\ref{tb:lightgbm-regions}. Function BeforeFindBestSplit() mainly spends its time on generating three arrays: the indices of training samples on this leaf, and the corresponding hessian and gradient values for them. Its time complexity is $O(N_r)$, where $N_r$ is the number of samples on the current leaf. Function ConstructHistogram() implements Algorithm 1, which goes over one feature data to construct its feature histogram in $O(N_r)$ time, and there are $d$ calls to this function (one for each feature), so the total complexity is $O(N_r d)$. Function FindBestThreshold() goes over one feature histogram to find the best split point with complexity $O(k)$, and there are $d$ calls to this function so the overall complexity is $O(kd)$. 
Function Split() is to split the samples on a node to its left and right children given
the split value and the feature index computed by the previous function. Its complexity 
is $O(N_r)$. 

As shown in Table~\ref{tb:lightgbm-regions}, on 3 different datasets with different $d$, $n$ and we fix $k=255$, the majority of time in LightGBM is spent on constructing histograms for all features in function ConstructHistogram(), which is expected, as this process needs to go over all feature values on one leaf ($O(N_r d)$ elements) to build $d$ histograms, each with $k$ bins. Since $k$ (usually 255 or smaller) is much smaller than $N_r$, the time spent on finding the best split inside histograms is not significant.

\begin{table}[htbp]
\centering
 \begin{tabular}{| c | c | c | c |} 
 \hline
 Function Name & \higgs & \dataepsilon & \yahooltr \\ [0.5ex] 
 \hline
 BeforeFindBestSplit() & 5.3\% & $<$ 0.1\% & 0.3\%\\
 \hline
 ConstructHistogram() & 80.8\% & 87.6\% & 75.7\%  \\ 
 \hline
 FindBestThreshold() & 0.4\% & 5.10\% & 6.80\% \\
 \hline
 Split() & 6.3\% & $<$ 0.1\% & 4.30\% \\
 \hline
\end{tabular}
\caption{Percentage of time for 4 most time-consuming functions in LightGBM.}
\label{tb:lightgbm-regions}
\end{table}

Our goal is thus to improve the performance of function ConstructHistogram() on GPU. It is clear that this function is memory-bound, as building the histograms requires non-sequential scattering access to large arrays in memory, in which case the cache system does not work quite well. Thus, our key to success is to utilize the high bandwidth memory and large computation power on GPUs to accelerate this operation.

\subsection{Synthetic benchmark on GPU Feature Histogram Construction}
\label{sec:synthetic}

We first test the speedup for building feature histograms on synthetic data. To construct synthetic data, we generate $n=8,000,000$ samples and each sample is $d=500$ dimensional. Each feature value is a byte holding a random bin number ranging from 1 to $k$, where $k \in \{64, 256\}$ is the total number of bins.

During the decision tree building process, each leaf node usually only holds a small portion of data samples. Considering a tree of depth $D$, one leaf will on average has $n/2^D$ training samples, and these training samples can spread far away in memory. Building histograms for a leaf node requires scanning data with a large, random stride in memory, which can be very slow because it is impossible to fit the whole dataset in cache. We emulate this behavior in our benchmark by generating a permutation of $n$ numbers, truncating it into size of $n/2^D$, sorting the truncated array, and then using it as indices to access data samples.

\begin{figure}[htbp]
\center
\includegraphics[width=2.5in]{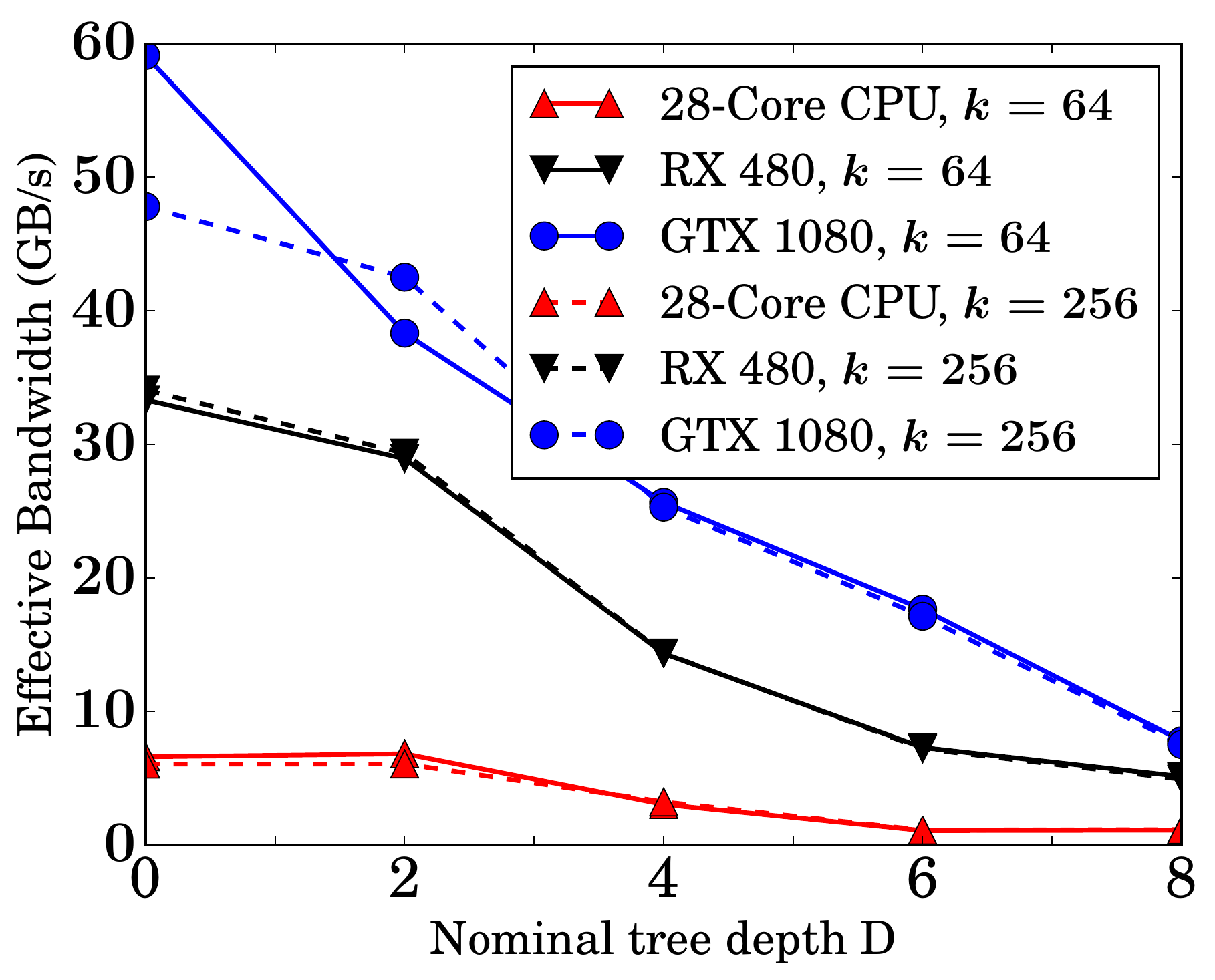}
\vspace{-1em}
\caption{Feature histogram building performance.}
\label{fig:benchmark}
\end{figure}

In Figure~\ref{fig:benchmark}, we compare the performance of our GPU algorithm with CPU for building feature histograms with $k=64$ and $256$. The CPU implementation is directly extracted from LightGBM's source code. The metric we used for comparison is the effective bandwidth; that is, how much feature data can be processed during one unit time. We observe that the largest bandwidth occurs when $D=0$. In this case, all the data are used to build histograms (e.g., histogram for the root of a tree), and memory access is sequential. When $D$ increases, processing bandwidth is reduced dramatically on all devices because of the large-stride memory access. GPU is much faster than CPU over all $D$s especially when $k=64$. However, when $D$ becomes too large, there are not many samples left, so the overhead of invoking a GPU function becomes significant, which limits the available speedup.

\subsection{Convergence Behaviour of Our GPU Histogram Algorithm}

\begin{figure}%
    \centering
    \subfloat[\higgs AUC vs Iterations]{{\includegraphics[width=2in]{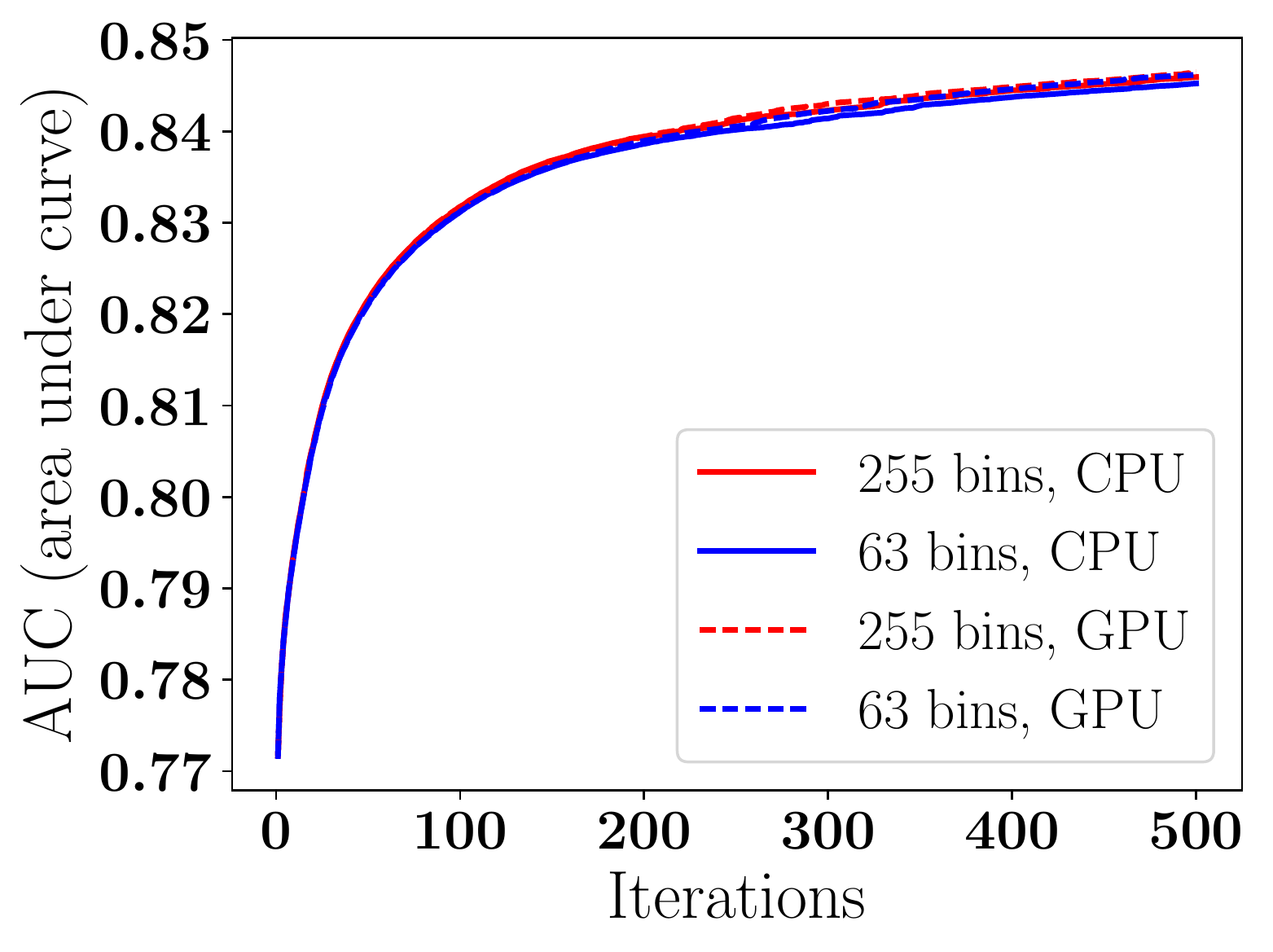}}}%
    \qquad
    \subfloat[\msltr NDCG@10 vs Iterations]{{\includegraphics[width=2in]{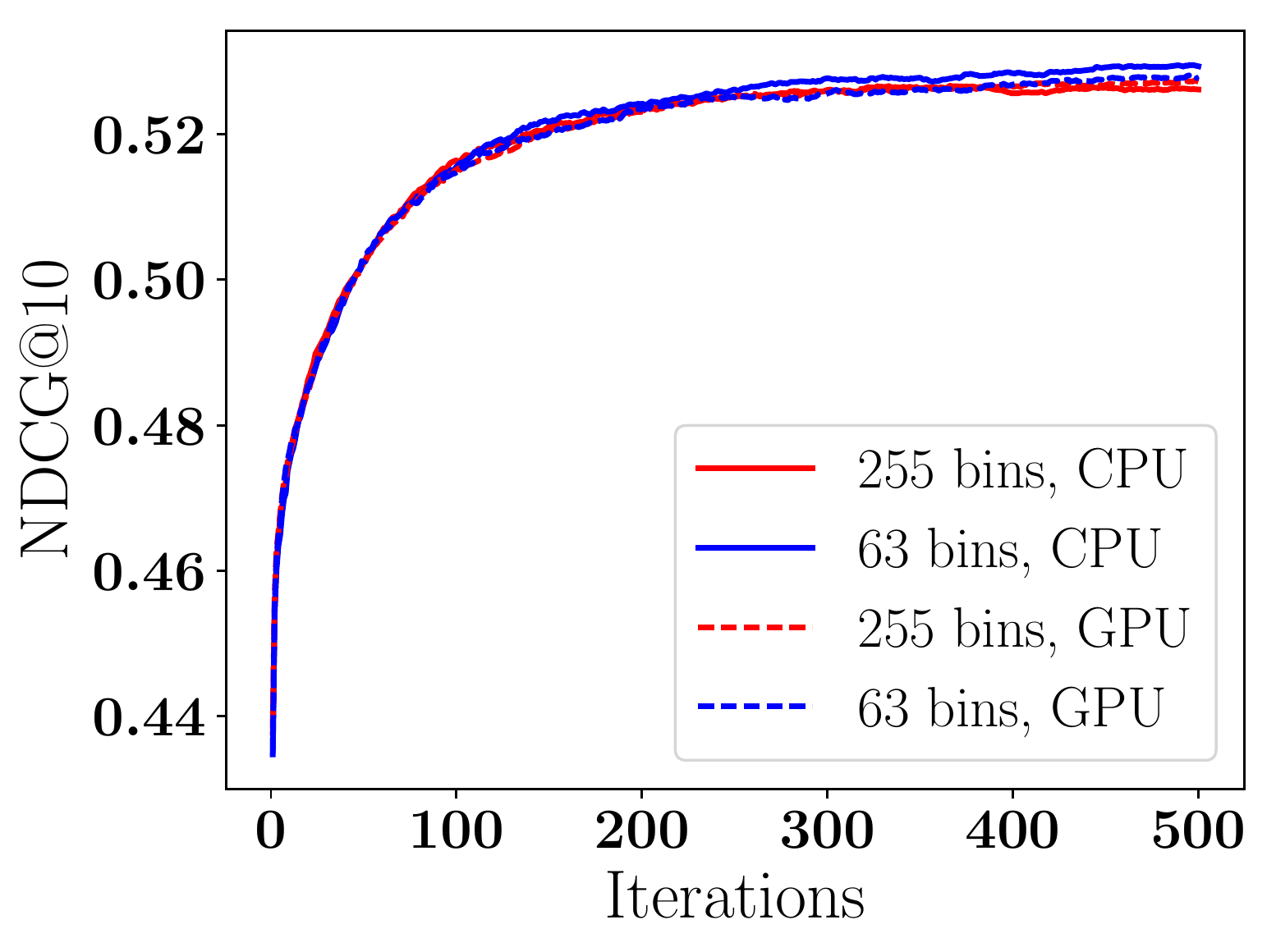}}}%
    \caption{Convergence Behaviour}%
    \label{fig:higgsandmsltr}
\end{figure}

Since our GPU algorithm uses a smaller bin size and single precision for training, we want to make sure that there is no instability during training. In Figure~\ref{fig:higgsandmsltr}, we show the AUC or NDCG@10 with respect to the number of boosting iterations for \higgs and \msltr. As we can see, despite of the different bin sizes, {\sf Histogram} on CPU (with double precision math) and GPU (with single precision math) both converge to the same metric value. Other datasets also exhibit a similar behaviour, so we omit their figures. 

\subsection{Use 4-bit Bins ($k=16$)}

It is possible to just use 4 bits to store a binned feature value (i.e., two feature values packed into one byte), by using a bin size of 16 (practically 15 bins in LightGBM as one bin is used as a sentinel). The benefits of using a 4-bit bin is three-fold: First, this reduces memory usage for storing feature values by half and also reduces required memory bandwidth; second, this allows 8 features to be packed into one 4-byte tuple, increasing the available workload per workgroup; third, a smaller bin size also reduces memory pressure in local memory, and allows multiple banks of histogram counters to be built to further reduce atomic update conflicts.

\begin{figure*}[htbp]
\hspace{-2em}\includegraphics[width=0.666\paperwidth]{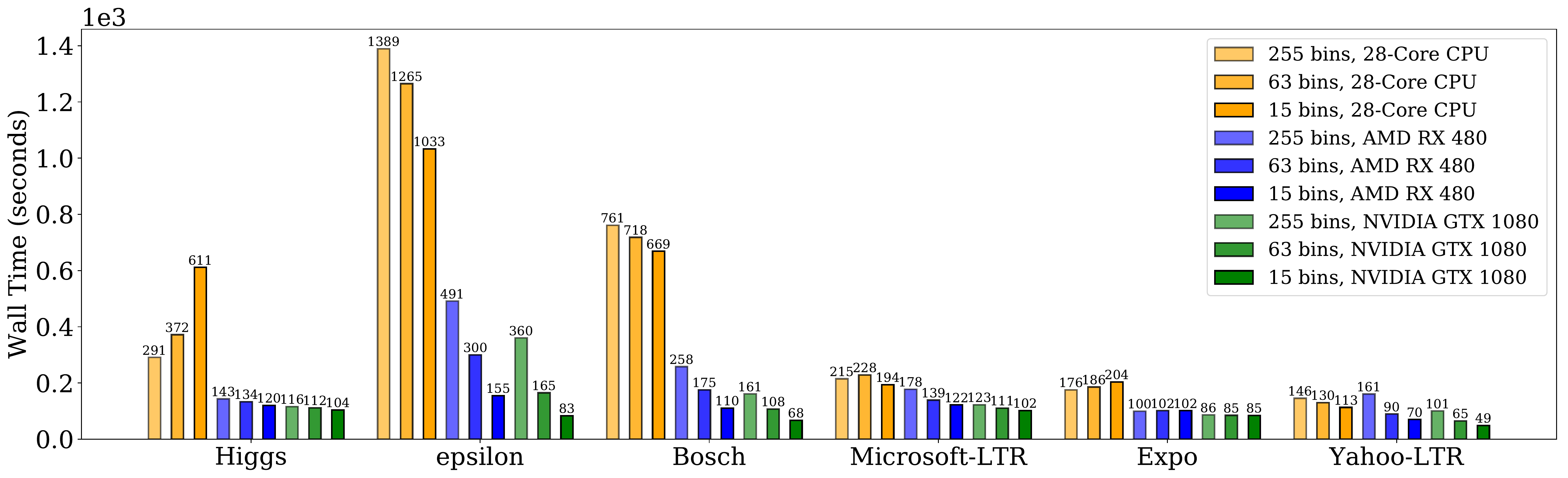}
\caption{Performance comparison of {\sf Histogram} algorithm with $k=\{15,63,255\}$ on CPU and GPU}
\label{fig:4bit-compare}
\end{figure*}

We conduct experiments for the same six datasets with the same settings as in the Experimental Results section, and compare training accuracy and speed in Table~\ref{tb:4bit-train-acc} and Figure~\ref{fig:4bit-compare}.
We can observe that although using 4-bit bins significantly decreases training time for some datasets (like \yahooltr, \dataepsilon and \bosch), we cannot achieve the same accuracy as using larger bin sizes with the same number of boosting iterations. However, sometimes the difference is very small, like in \higgs, \dataepsilon, \yahooltr and \bosch; thus, using a bin size of 15 may help us produce a reasonably good model within a very short time.

\begin{table*}[htbp]
\resizebox{14.5cm}{!}
{
\begin{tabular}{|l|l|c|c|c|c|c|c|}
\hline
Dataset                    & Metrics & \begin{tabular}[c]{@{}c@{}}CPU\\ $k=255$\end{tabular} & \begin{tabular}[c]{@{}c@{}}CPU\\ $k=63$\end{tabular} & \begin{tabular}[c]{@{}c@{}}CPU\\ $k=15$\end{tabular} & \begin{tabular}[c]{@{}c@{}}GPU\\ $k=255$\end{tabular} & \begin{tabular}[c]{@{}c@{}}GPU\\ $k=63$\end{tabular} & \begin{tabular}[c]{@{}c@{}}GPU\\ $k=15$\end{tabular} \\ \hline
\higgs                      & AUC     & 0.845612                                              & 0.845239                                             & 0.841066                                             & 0.845612                                              & 0.845209                                             & 0.840748                                             \\ \hline
\dataepsilon                    & AUC     & 0.950243                                              & 0.949952                                             & 0.948365                                             & 0.950057                                              & 0.949876                                             & 0.948365                                             \\ \hline
\multirow{4}{*}{\yahooltr} & NDCG@1  & 0.730824                                              & 0.730165                                             & 0.729647                                             & 0.730936                                              & 0.732257                                             & 0.73114                                              \\ \cline{2-8} 
                           & NDCG@3  & 0.738687                                              & 0.737243                                             & 0.736445                                             & 0.73698                                               & 0.739474                                             & 0.735868                                             \\ \cline{2-8} 
                           & NDCG@5  & 0.756609                                              & 0.755729                                             & 0.754607                                             & 0.756206                                              & 0.757007                                             & 0.754203                                             \\ \cline{2-8} 
                           & NDCG@10 & 0.79655                                               & 0.795827                                             & 0.795273                                             & 0.795894                                              & 0.797302                                             & 0.795584                                             \\ \hline
\expo                       & AUC     & 0.776217                                              & 0.771566                                             & 0.743329                                             & 0.776285                                              & 0.77098                                              & 0.744078                                             \\ \hline
\multirow{4}{*}{\msltr}    & NDCG@1  & 0.521265                                              & 0.521392                                             & 0.518653                                             & 0.521789                                              & 0.522163                                             & 0.516388                                             \\ \cline{2-8} 
                           & NDCG@3  & 0.503153                                              & 0.505753                                             & 0.501697                                             & 0.503886                                              & 0.504089                                             & 0.501691                                             \\ \cline{2-8} 
                           & NDCG@5  & 0.509236                                              & 0.510391                                             & 0.507193                                             & 0.509861                                              & 0.510095                                             & 0.50663                                              \\ \cline{2-8} 
                           & NDCG@10 & \multicolumn{1}{l|}{0.527835}                         & \multicolumn{1}{l|}{0.527304}                        & \multicolumn{1}{l|}{0.524603}                        & \multicolumn{1}{l|}{0.528009}                         & \multicolumn{1}{l|}{0.527059}                        & \multicolumn{1}{l|}{0.524722}                        \\ \hline
\bosch                      & AUC     & \multicolumn{1}{l|}{0.718115}                         & \multicolumn{1}{l|}{0.721791}                        & \multicolumn{1}{l|}{0.716677}                        & \multicolumn{1}{l|}{0.717184}                         & \multicolumn{1}{l|}{0.724761}                        & \multicolumn{1}{l|}{0.717005}                        \\ \hline
\end{tabular}
}
\caption{Training metrics comparison of {\sf Histogram} algorithm with $k=\{15,63,255\}$ on CPU and GPU
}
\label{tb:4bit-train-acc}
\end{table*}


\end{document}